# Multi-View Large-Scale Bundle Adjustment Method for High-Resolution Satellite Images

**Xu Huang** [a], Research Scholar
**Rongjun Qin** [a,b], Professor
[a] Department of Civil, Environmental and Geodetic Engineering, The Ohio State University,
218B Bolz Hall, 2036 Neil Avenue, Columbus, OH 43210, USA;
[b] Department of Electrical and Computer Engineering, The Ohio State University,
205 Dreese Labs, 2015 Neil Avenue, Columbus, OH 43210, USA
huang.3651@osu.edu
qin.324@osu.edu

**ABSTRACT**

Given enough multi-view image corresponding points (also called tie points) and ground control points (GCP), bundle adjustment for high-resolution satellite images is used to refine the orientations or most often used geometric parameters Rational Polynomial Coefficients (RPC) of each satellite image in a unified geodetic framework, which is very critical in many photogrammetry and computer vision applications. However, the growing number of high resolution spaceborne optical sensors has brought two challenges to the bundle adjustment: 1) images come from different satellite cameras may have different imaging dates, viewing angles, resolutions, etc., thus resulting in geometric and radiometric distortions in the bundle adjustment; 2) The large-scale mapping area always corresponds to vast number of bundle adjustment corrections (including RPC bias and object space point coordinates). Due to the limitation of computer memory, it is hard to refine all corrections at the same time. Hence, how to efficiently realize the bundle adjustment in large-scale regions is very important. This paper particularly addresses the multi-view large-scale bundle adjustment problem by two steps: 1) to get robust tie points among different satellite images, we design a multi-view, multi-source tie point matching algorithm based on plane rectification and epipolar constraints, which is able to compensate geometric and local nonlinear radiometric distortions among satellite datasets, and 2) to solve dozens of thousands or even millions of variables bundle adjustment corrections in the large scale bundle adjustment, we use an efficient solution with only a little computer memory. Experiments on in-track and off-track satellite datasets show that the proposed method is capable of computing sub-pixel accuracy bundle adjustment results.

## 1. INTRODUCTION

With the development of high-resolution (high-res) satellite imaging technology, e.g. World View-1/2/3, GeoEye, Quickbird, IKONOS, Gaofen-2, Jilin-1, Gaojing-1, etc., every place of the earth can be covered by dozens or even hundreds of high-res satellite images with different imaging date. It is promising to utilize these high-res satellite images in 3D reconstruction, semantic classification, monitoring tracking, change detection, etc. However, the absolute geo-referencing accuracy (in 90% circular error metric) of these high-res satellite images are not satisfactory, e.g. World View-1 with accuracy: 4.0m (DigitalGlobe, 2016a), GeoEye with accuracy: 3.0m (DigitalGlobe, 2016b), Quickbird with accuracy: 23m (DigitalGlobe, 2016c). The geo-referencing errors must be corrected before utilizing these high-res satellite images in the practical applications.

All these high-res satellite images provide Rational Polynomial Coefficient (RPC) camera model which is a fractional polynomial with 80 parameters (two parameters are normally set as one), describing the geometric relationships between object space points (normally defined as latitude, longitude, elevation) and image pixels (normally defined as row, column). The flying height of the satellites is typically around 500 km, making the optical rays of each pixels are almost parallel to each other. Therefore, the geo-referencing errors can be corrected by small translations in the image plane, which is also called bias correction (Grodecki and Dial, 2003).

Given enough tie points and optionally available ground control points (GCP), the bias as well as the ground coordinates of these tie points can be corrected through bundle adjustment techniques. Some previous work (d'Angelo and Reinartz, 2012; Fraser and Hanley, 2005; Hanley and Yamakawa, 2002; Jacobsen, 2008; Xiaohua et al., 2010) focused on correcting bias among the images from the same satellite cameras. However, these methods were not fit for large-scale bundle adjustment for two reasons: 1) in large scale scenarios, it is difficult to find enough images from the same satellites for the bias correction. Thus, satellite images from different satellite cameras are good supplements in this case, while these images often have different imaging time, viewing angle, and ground sampling distances (GSD), thus may bring geometric and radiometric distortions to tie point matching; 2) large scale bundle adjustment may bring dozens of thousands or even millions of variables (mostly corrections of object space point coordinates and small number of bias corrections), making it difficult to simultaneously solve such a large quantity of variables. An alternative solution is to divide satellite images into groups, run the bundle adjustment independently on each group, and then registering these group one by one (Ozcanli et al., 2014), while the registering will bring propagation errors to the final geo-referencing accuracies.

This paper particularly addresses the above multi-view large-scale bundle adjustment problem by two steps: 1) to get robust tie points among different satellite images, we design a multi-view, multi-source tie point matching algorithm based on plane rectification and epipolar constraints, and 2) to solve dozens of thousands or even millions of variables (including image bias corrections and ground coordinate corrections) in the large scale bundle adjustment, we use an efficient solution with only a little computer memory. For 1), we firstly rectified all satellite images onto a same height plane (also called level-2 image correction) to highly reduce the geometric distortions (e.g. rotation, translation, affine distortion, scaling) among multi-view images, thus the bias among images can be simplified as constants (Morgan, et al., 2004); then, we select image pairs from the level-2 products based on their overlaps to avoid unnecessary blind matching; after image selection, FAST detector (Viswanathan, 2009) is used to detect features on level-2 products, and their correspondences are found through our proposed multi-block census (MBCensus) matching metric along epipolar constraints, which divides matching window into 9 blocks, thus being able to describing complicated object structures and compensate local nonlinear radiometric distortions between pairs. The workflow of our algorithm is shown in Figure 1.

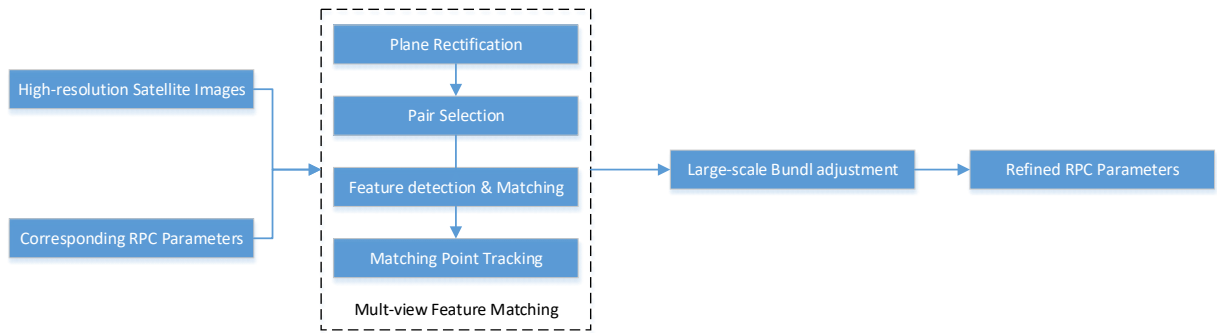

Figure 1. Workflow of our algorithm

The rest of the paper is organized as follows: Section 2 gives a brief introduction of RPC model and the corresponding constant bias modeled RPC; Section 3 introduces our multi-view, multi-source feature matching algorithm; Section 4 introduces an efficient solution to large scale bundle adjustment method; Section 5 shows the experimental results; and Section 6 draws the conclusions based on our works.

## 2. BIAS MODELED RPC

RPC model maps object space points (latitude, longitude, height) onto image points (row and column coordinates) through one fractional polynomial function with 80 parameters. To minimize numerical errors in the calculation, normalized image coordinates and ground coordinates are actually used in RPC model:

$$r = (row - LINE\_OFF)/LINE\_SCALE \quad c = (col - SAMP\_OFF)/SAMP\_SCALE$$
$$P = (lat - LAT\_OFF)/LAT\_SCALE$$
$$L = (long - LONG\_OFF)/LONG\_SCALE \quad (1)$$
$$H = (hei - HEI\_OFF)/HEI\_SCALE$$

where, $row$ and $col$ are image coordinates in row and column directions; $LINE\_OFF$, $SAMP\_OFF$ are offset values for image coordinates; $LINE\_SCALE$, $SAMP\_SCALE$ are scale values for image coordinates; $lat$, $long$, $hei$ are ground coordinates of latitude, longitude and height; $LAT\_OFF$, $LONG\_OFF$ and $HEI\_OFF$ are offset values for ground coordinates; $LAT\_SCALE$, $LONG\_SCALE$ and $HEI\_SCALE$ are scale values for ground coordinates; $r$, $c$ are normalized image coordinates in row and column directions; $P$, $L$, $H$ are normalized ground coordinates of latitude, longitude and height.

After the normalization, each polynomial in RPC model is up to third order in ($P$, $L$, $H$) with 20 terms, and the RPC model can be expressed as:

$$r = \frac{\sum_{i=1}^{20} LINE\_NUM\_COEF_i \cdot p_i(P,L,H)}{\sum_{i=1}^{20} LINE\_DEN\_COEF_i \cdot p_i(P,L,H)} \quad c = \frac{\sum_{i=1}^{20} SAMP\_NUM\_COEF_i \cdot p_i(P,L,H)}{\sum_{i=1}^{20} SAMP\_DEN\_COEF_i \cdot p_i(P,L,H)} \quad (2)$$

where, $LINE\_NUM\_COEF_i$ ($i$ = 1…20), $LINE\_DEN\_COEF_i$ ($i$ = 1…20), $SAMP\_NUM\_COEF_i$ ($i$ = 1…20) and $LINE\_DEN\_COEF_i$ ($i$ = 1…20) are 80 parameters of RPC models; $p_i(P,L,H)$ is a three-order function with variables $P$, $L$, $H$, Detailed description of $p_i(P,L,H)$ can be referred to Quickbird product guide (DigitalGlobe, 2016d).

Bias between image pairs can be formulated as small translations in image space, as follows:

$$\begin{aligned} f_{row} &= row + \Delta row - r \cdot LINE\_SCALE - LINE\_OFF \\ f_{col} &= col + \Delta col - c \cdot SAMP\_SCALE - SAMP\_OFF \end{aligned} \quad (3)$$

where, $\Delta row$ and $\Delta col$ are bias in row and column directions; $f_{row}$ and $f_{col}$ are two bias modeled RPC functions in row and column directions. The goal of our work is to compute $\Delta row$ and $\Delta col$ for each image. It is well known that a simple constant bias ($\Delta row, \Delta col$) may not be able to compensate non-linear distortions (caused by inconsistent platform speed and rotations) in the original satellite images. Therefore, before the bias correction, a plane rectification (section 3.1) is needed that project the satellite images to a common height plane to eliminate the non-linear distortions.

## 3. MULTI-VIEW, MULTI-SOURCE FEATURE MATCHING

High-res satellite images from different satellite cameras will result in different scaling, radiometric conditions and even geometric shapes, thus bringing several uncertainties in tie point matching. Since the bias between image pairs are often corrected by tie points, the matching accuracies directly determine the bias correction accuracy. Therefore, our work firstly focuses on how to obtain robust and accurate tie points in multi-view, multi-source scenarios. In this paper, we formulate the multi-view feature matching problem into multi-pair matching. To reduce the geometric distortions (e.g. scaling, rotation, etc.), all images are rectified in the same height plane with the same GSD resolution (in section 3.1); then, image pairs with enough overlaps are selected to avoid unnecessary blind matching (in section 3.2); For each image pair, FAST detector (Viswanathan, 2009) is utilized to detect corner features, and the correspondences are matched by our proposed multi-block Census metric along epipolar constraints, which is robust to local nonlinear radiometric distortions; finally, multi-view point tracking is conducted to find correspondences in multi-view images (section 3.4).

### 3.1 Plane Rectification

Plane rectification (Toutin, 2004) is a process to map images onto a common height plane in a unified geodetic system, also termed as level-2 image correction. The sampling GSD of each level-2 products can be fixed, and the geo-referencing of them are similar, thus greatly reducing geometric distortions between pairs, as shown in Figure 2. The first row in Figure 2(a) are original images with scaling and affine distortions, and the second row in Figure 2(b) are rectified images with similar geometric shapes. In our work, the common height plane is obtained from rpc files by averaging the height offsets $HEI\_OFF$. The GSD of level-2 products is computed by selecting maximum GSD of each original satellite image which is computed by projecting these images on the common height plane, and then dividing the projecting area by the pixels number of each image. After then, RPC parameters for each level-2 products need to be generated. In this paper, we used the RPC of original images to generate several virtual GCPs, and utilize these GCPs to compute RPC for each level-2 products (Fraser and Dial, 2006).

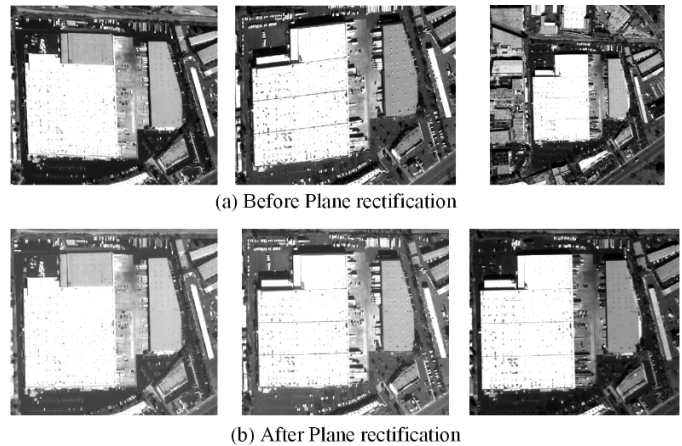
(a) Before Plane rectification

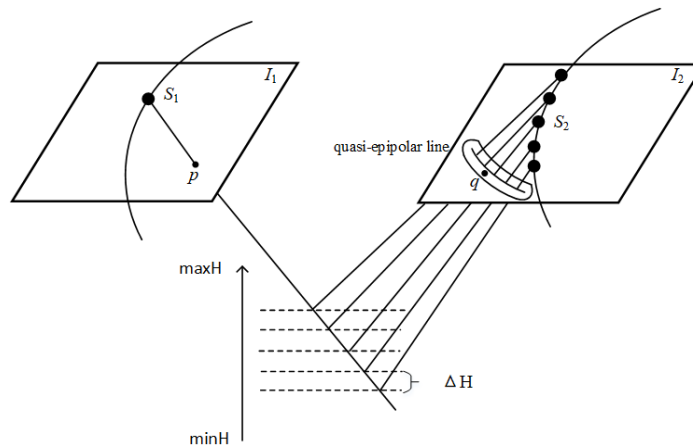
(b) After Plane rectification

Figure 2. An example of plane rectification. The images from left to right in (a) are two World View 3 images with GSD 0.3 and a World View 2 image with GSD 0.5m; The images in (b) are corresponding rectified images with GSD 0.5m.

### 3.2 Pair Selection

The pair selection strategy in this paper is based on overlap between each pair of level-2 products. Since each level-2 products already have geo-location information for each pixel, we can compute the bounding box of level-2 products in ground space, and the overlap of each pair is computed by the intersection of the bounding boxes. We select image pairs whose division between the overlap and the smaller bounding box is larger than a predefined threshold (0.6 in this paper).

### 3.3 Feature Detection and Matching

In this paper, we utilize FAST detector to detect corner features on each level-2 products. It is well known that the correspondences of these features must lie on quasi-epipolar lines in high-res linear-array images (Lee et al., 2003). However, the bias between these pairs will make the quasi-epipolar line dozens of pixels away from its true positions. To compensate these bias, we search correspondences of each feature in the buffers centered at the epipolar lines, as shown in Figure 3. $\{I_1, I_2\}$ is a pair of level-2 products with $I_1$ being left image and $I_1$ being right image, $\{S_1, S_2\}$ are two satellite cameras; $\{p, q\}$ is a pair of correspondences. The ray from $p$ is cast onto a series of height planes with elevations in the range [minH, maxH] and $\Delta$H increments, and the intersections between the ray and the height planes are back-projected on the other images, thus generating a quasi-epipolar line. Then, a searching buffer is centered at the epipolar line with a certain radius (30 pixels in this paper), and the correspondence is found by compare the appearance similarity between $p$ and the corner features in the searching buffer.

Figure 3. Epipolar constrained Feature Matching.

Census is an often used matching cost metric which transform matching windows into binary strings, and compares the hamming distance between them (Zabih and Woodfill, 1994). It is fast and robust in nonlinear radiometric distortions, while it is sensitive to noise and failed to describe complex object structures. In this paper, we improved traditional census metric by applying Gaussian filtering to remove noises and then dividing the matching window into 9 blocks to better describe complex structures, thus termed as multi-block census, as shown in Figure 4. Each block can provide a binary string, and the feature descriptor of the matching window in the left and right image is consist of the binary strings of the 9 blocks. The final matching score is computed by comparing the hamming distances of the feature descriptors in Equation (4).

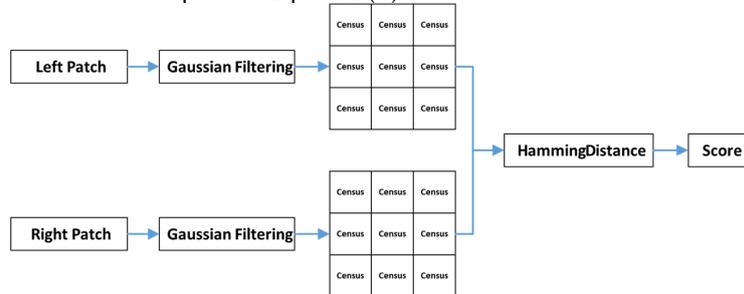

Figure 4. MBCensus matching metric

$$C_{MBCensus}(p,q) = ||F_{I_1}(p,\{Census_i\}) - F_{I_2}(q,\{Census_i\})||_H \quad i = 1, \ldots, 9 \qquad (4)$$

where, $Census_i$ is the census binary string of the $i$ th block, $F_{I_1}$ and $F_{I_2}$ are MBCensus feature descriptors in the pair $\{I_1, I_2\}$, $|| \;||_H$ means the hamming distance, $C_{MBCensus}$ is the matching score of MBCensus. Lower $C_{MBCensus}$ means higher possibility of being correspondences.

Scale-invariant feature transform (SIFT) (Lowe, 2004) matching metric is perhaps the most often used image matching algorithm which can get robust matching points with compensating some geometric distortions (scaling and rotation) and linear radiometric distortions. However, in the case of satellite image matching, the geometric distortions can be rectified by plane rectification in section 3.1, and due to sun elevation angles, seasonal changes and object materials, there often exist nonlinear radiometric distortions in satellite images, which may reduce the matching accuracies of SIFT. To give a comparison between SIFT and MBCensus, we selected a stereo pair in San Diego with different imaging time, as shown in Figure 5.

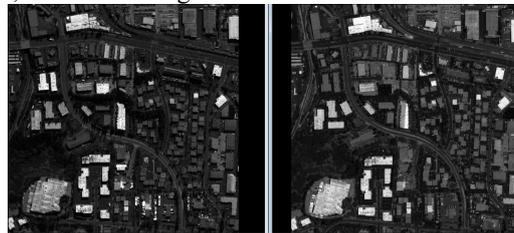

Figure 5. Stereo pair in San Diego

We realized SIFT matching and MBCensus matching in the pair, and compared them by three aspects: 1) correspondences numbers; 2) average reprojection errors after bias correction using these correspondences; 3) average matching time for each pixel. The comparison result is shown in Table 1. It shows that MBCensus can get lower matching time and comparable matching accuracies, which is suitable for large-scale satellite image matching.

Table 1. Comparison between SIFT and MBCensus

|  | SIFT | MBCensus |
| --- | --- | --- |
| Correspondences Number | 7803 | 7437 |
| average reprojection errors (pixel) | 0.330 | 0.161 |
| average matching time (ms) / pixel | 5.13 | 3.59 |

To get more robust matching results, we detect and eliminate mismatches by two strategies: 1) For each feature point in left image, we record the two potential correspondences on right image with lowest and the second lowest MBcensus score, and choose the one with the ratio between the two scores lower than 0.6; 2) we compensate bias between correspondences and their corresponding epipolar line, and use the bias corrected RPC model to compute the object space points from correspondences, which is then back-projected onto images to measure the reprojection error (the distance between projected image points and the original points). We eliminate the correspondences with reprojection error larger than 2 pixels.

### 3.4 Multi-view Point Tracking

After feature matching, there must be some correspondences in one image pair corresponding to the same object space point with correspondences in other image pairs. Multi-view point tracking is to find feature points in multi-view images that are corresponding to a same object space point, which is critical important in eliminating correlations among object space points in the large scale bundle adjustment in section 4. In this paper, we track points by checking if correspondences in different image pairs have common feature points.

## 4. LARGE SCALE BUNDLE ADJUSTMENT

Large scale bundle adjustment is to globally compute bias among multi-view satellite images (Qin, 2016). Given good initial bias values (normally set as zero), the nonlinear bias corrected RPC model in Equation (3) can be linearized by Taylor expansion, and then we can solve the optimal solutions of bias iteratively. In this part, we firstly introduce the object energy function of the large-scale bundle adjustment in section 4.1, and then introduce how to get the optimal solution of bias based on an efficient solution in section 4.2.

### 4.1 Object energy function

Large scale bundle adjustment is to globally compute bias among multi-view satellite images such that the reprojections errors of each object space points can be minimized. If ground control point (GCP) is available, the object space point coordinates after bundle adjustment should also be close to its corresponding GCP. Thus, the final object energy function can be formulated as:

$$\min E(\Delta row, \Delta col, lat, lon, hei)$$
$$= \sum_i \sum_j ||RPC(\Delta row_i, \Delta col_i, lat_j, lon_j, hei_j) - p_j^i||$$
$$+ \sum_{j \in GCP} ||(lat_j, lon_j, hei_j)^T - (lat_{j,g}, lon_{j,g}, hei_{j,g})^T|| \quad (5)$$

where, $\Delta row = \{\Delta row_i\}$, $\Delta col = \{\Delta col_i\}$ are set of image bias; $lat = \{lat_j\}$, $lon = \{lon_j\}$ and $hei = \{hei_j\}$ are set of space object point coordinates; RPC means biased RPC model in Equation (3); $p_j^i$ is the image pixel of $j$ th object space points in $i$ th image; $(lat_{j,g}, lon_{j,g}, hei_{j,g})$ are ground coordinates of GCP.

### 4.2 Global bundle adjustment

Given initial values of image bias (zero in this paper) and object space point coordinates (computed by coarse RPC parameters of each image), we solve the optimal solution of the energy object function by least square method with Taylor expansion. Set $X_i = (d\Delta row_i, d\Delta col_i)^T$ as the bias corrections in the $i$ th image with $d\Delta row$ being corrections in row direction and $d\Delta col$ being the corrections in column direction, $Y_j = (dLat_j, dLon_j, dHei_j)^T$ as the ground coordinate corrections of the $j$ th object space points with $dLat$, $dLon$ and $dHei$ being the corrections of latitude, longitude and height, $b_i^0 = (\Delta row_i^0, \Delta col_i^0)^T$ as the initial values of bias in $i$ th images with $\Delta row_i^0$ being initial values in row directions and $\Delta col_i^0$ being initial values in column directions, $g_j^0 = (Lat_j^0, Lon_j^0, Hei_j^0)^T$ as the initial values of ground coordinates of the $j$ th points with $Lat_j^0$, $Lon_j^0$ and $Hei_j^0$ being initial values of latitude, longitude and height, the 1-order Taylor expansion of Equation (3) can be expressed as:

$$V_{i,j} = \begin{pmatrix} V_{row}(i,j) \\ V_{col}(i,j) \end{pmatrix}$$
$$= \begin{pmatrix} \frac{\partial f_{row_{i,j}}(b_i^0, g_j^0)}{\partial \Delta row_i} & 0 \\ 0 & \frac{\partial f_{col_{i,j}}(b_i^0, g_j^0)}{\partial \Delta col_i} \end{pmatrix} \begin{pmatrix} d\Delta row_i \\ d\Delta col_i \end{pmatrix} + \begin{pmatrix} \frac{\partial f_{row_{i,j}}(b_i^0, g_j^0)}{\partial Lat_j} & \frac{\partial f_{row_{i,j}}(b_i^0, g_j^0)}{\partial Lon_j} & \frac{\partial f_{row_{i,j}}(b_i^0, g_j^0)}{\partial Hei_j} \\ \frac{\partial f_{col_{i,j}}(b_i^0, g_j^0)}{\partial Lat_j} & \frac{\partial f_{col_{i,j}}(b_i^0, g_j^0)}{\partial Lon_j} & \frac{\partial f_{col_{i,j}}(b_i^0, g_j^0)}{\partial Hei_j} \end{pmatrix} \begin{pmatrix} dLat_j \\ dLon_j \\ dHei_j \end{pmatrix} - \begin{pmatrix} -f_{row_{i,j}}(b_i^0, g_j^0) \\ -f_{col_{i,j}}(b_i^0, g_j^0) \end{pmatrix} \quad (6)$$

where, $V_{i,j}$ is the residual error of the $j$ th object space point on the $i$ th image, which is actually related to the distances between original image pixels and its projected pixels from the corresponding object space points; $V_{row}$ is the residual error in row direction; $V_{col}$ is the residual error in the column direction.

Using $A_i$ as the coefficient matrix of the first term in Equation (6), $B_j$ as the coefficient matrix of the second term and $L_{i,j}$ as the constant vector of the second term, Equation (6) can be simplified as:

$$V_{i,j} = A_i \cdot X_i + B_j \cdot Y_j - L_{i,j} \tag{7}$$

If GCPs are given, the ground coordinates of GCPs must be fixed during bundle adjustment. Therefore, the ground coordinate corrections of GCPs must be zero, as follows:

$$Y_g = 0 \tag{8}$$

where, $Y_g$ is the ground coordinate corrections of GCPs.

Given $N$ images with $T$ image points and the corresponding $M$ object space points (containing $m$ GCPs), we will have $2 \times N$ variables of bias corrections and $3 \times M$ variables of object space point coordinates corrections. Set $X = \{X_0, X_1, ..., X_N\}$ as the sets of bias corrections; $Y = \{Y_0, Y_1, ..., Y_M\}$ as the sets of object space point coordinate corrections, similar to Equation (7) and (8), the whole bundle adjustment functions can be expressed as:

$$V = \begin{pmatrix} A & B \\ 0 & G \end{pmatrix} \begin{pmatrix} X \\ Y \end{pmatrix} - \begin{pmatrix} L \\ 0 \end{pmatrix} = CU - D \tag{9}$$

where, $V$ is the residual error vectors of all image points; $A$ is a $2T \times 2N$ sparse coefficient matrix consisted of $A_i$ ($i = 1, 2, ..., N$); $B$ is a $2T \times 3M$ sparse coefficient matrix consisted of $B_j$ ($j = 1, 2, ..., M$); $G$ is a $3m \times 3M$ sparse coefficient matrix related to GCPs; $L$ is a $2T \times 1$ vector consisted of constant term $L_{i,j}$; $X$ is a bias correction vector consisted of $X_i$, and $Y$ is a ground coordinate correction vector consisted of $Y_i$; $C$ is a $(2T + 3m) \times (2N + 3M)$ sparse coefficient matrix of the whole bundle adjustment functions; $D$ is a $(2T + 3m) \times 1$ vector representing constant terms of the whole bundle adjustment functions; $U$ is the unknown vector consisted of all bias corrections and ground coordinate corrections.

Since the number of images (up to thousands) are much smaller than the number of object space points (from dozens of thousands to millions), the size of coefficient matrix $C$ mainly dependent on the number of object space points. The number of object space points are so large that it is difficult to simultaneously solve all corrections. The traditional way to solve this problem is to eliminate the large quantity number of object space point coordinate corrections in the bundle adjustment functions, as shown in Equation (10), thus only bias corrections are computed. The object space point coordinates can be corrected by new bias modeled RPC (Egels and Kasser, 2003).

$$\begin{aligned}[N_A - N_{AB} N_B^{-1} N_{AB}^T] X &= L_A - N_{AB} N_B^{-1} L_B \\ N_A = A^T A \quad N_B &= B^T B + G^T G \quad N_{AB} = A^T B \\ L_A &= A^T L \quad L_B = B^T L \end{aligned} \tag{10}$$

where, $N_A$ is a $2N \times 2N$ diagonal matrix with diagonal element being related to $A_i^T A_i$ and other elements being 0; $N_B$ is a $3M \times 3M$ diagonal matrix with diagonal element being realted to $B_j^T B_j$ and other elements being 0; $N_{AB}$ is a $2N \times 3M$ correlation matrix of $A$ and $B$; $L_A$ and $L_B$ are constant vectors consisted of $A$, $B$ and $L$.

Though large quantity of object space point coordinates can be eliminated in Equation (10), the large size of $N_B$ and $N_{AB}$ still make the solution of Equation (10) difficult to solve. In this paper, we go on simplifying Equation (10) by directly computing $N_{AB} N_B^{-1} N_{AB}^T$ and $N_{AB} N_B^{-1} L_B$ instead of respectively computing $N_B$, $N_{AB}$, $L_A$ and $L_B$. Since the number of images $N$ is only up to thousands, both sizes of $N_{AB} N_B^{-1} N_{AB}^T$ and $N_{AB} N_B^{-1} L_B$ are only $2N \times 2N$ and $2N \times 1$, which is appropriate for large-scale bundle adjustment.

In particular, each object space point is corresponding to several images such that $B_j$ is different on different images. We define $B_{i,j}$ as coefficient matrix of the $j$ th object space point on the $i$ th image, and $N_B$ can be expressed as $Diag(\sum_{i=1}^{N} \delta(i,j) \cdot B_{i,j}^T B_{i,j})$ ($j = 1,..., M$) with $Diag$ being diagonal matrix and $\delta(i,j)$ being a indicator function which is 1 if $j$ th point is visible on the $i$ th image, otherwise, 0. Its inverse matrix is $Diag\left(\sum_{i=1}^{N} \delta(i,j) \cdot \left(B_j^T B_j\right)^{-1}\right)$, and $N_{AB}$ can be expressed as the matrix consisted of $A_i^T B_j$ ($i = 1,...,N; j = 1,..., M$) with $A_i^T B_j$ being its sub-matrix in $i$ th row, $j$ th column. Therefore, $N_{AB} N_B^{-1} N_{AB}^T$ can be computed directly:

$$N_{AB} N_B^{-1} N_{AB}^T = [A_i^T \left(\sum_{j=1}^{M} \delta(i,j) \cdot \delta(k,j) \cdot B_j N_B(j)^{-1} B_j^T\right) A_k]_{i,k} \quad i, k = 1,...,N; \tag{11}$$

where, $[\ ]_{i,k}$ means a matrix with its element in $i$ th row and $j$ th column; $N_B(j)$ is the $3 \times 3$ sub-matrix of the $j$ th object space point along the diagonal of $N_B$. In Equation (11), both $A_i$ and $A_k$ are $2 \times 2$ matrix, $B_j$ is a $2 \times 3$ matrix and $N_B(j)$ is a $3 \times 3$ matrix, thus no large matrix will be built during the computation of $N_{AB}N_B^{-1}N_{AB}^T$.

Similar to Equation (11), $L_A$ can be represented as $L_A = [\sum_{j=1}^{M} \delta(i,j) \cdot A_i^T L_{i,j}]_{i,1}$ with $L_{i,j}$ being the constant term of $j$ th point on the $i$ th image, $L_B$ can be represented as $L_B = [\sum_{i=1}^{N} \delta(i,j) \cdot B_j^T L_{i,j}]_{j,1}$, and the $N_{AB}N_B^{-1}L_B$ can be represented as:

$$N_{AB}N_B^{-1}L_B = [A_i^T(\sum_{j=1}^{M} \delta(i,j) \cdot B_j N_B(j)^{-1} L_B(j))]_{i,1}\ i = 1,\dots,N; \quad (12)$$

where, $L_B(j)$ is the $j$ th sub-vector corresponding to the $j$ th point in $L_B$.

According to Equation (11) and (12), the maximum matrices in the bundle adjustment are $N_A$ and $N_{AB}N_B^{-1}N_{AB}^T$, both of which are only $2N \times 2N$. As image number is usually not much (up to thousands), the memory of $N_A$ and $N_{AB}N_B^{-1}N_{AB}^T$ are accepted in most compute configurations.

Finally, the solution of all bias corrections can be optimized through bundle adjustment technique:

$$X = [N_A - N_{AB}N_B^{-1}N_{AB}^T]^{-1}(L_A - N_{AB}N_B^{-1}L_B) \quad (13)$$

The initial values of bias $\Delta row_i, \Delta col_i$ ($i = 1,\dots,N$) can be updated by the bias corrections, and then utilize them in the next iteration. The iteration will stop when the average re-projection error changes between the adjacent iterations is smaller than a predefined threshold (0.001 pixel in this paper). It is worth noting that the GCPs in only optional in this paper. If no GCPs are available, the GCP related matrix $G = 0$, and the bundle adjustment in Equation (9) is actually the free network adjustment.

## 5. EXPERIMENTAL COMPARISION AND ANALYSIS

### 5.1 Data Description

To test our bundle adjustment method in different datasets (in track or off track), we selected five in-track images in London and five off-track images in San Diego. The London dataset were World-View 2 satellite images (0.5 m GSD), collected on October 22, 2011, covering 50 km$^2$ area, as shown in Figure 6(a). The San Diego dataset consisted of both World-View 2 (GSD: 0.5m) and World-View 3 dataset (GSD: 0.3m), collected from August 14, 2015 to February 06, 2016, covering 100 km$^2$ area, as shown in Figure 6(b). GCPs are not available in both datasets, so out experiments were based on the free network adjustment.

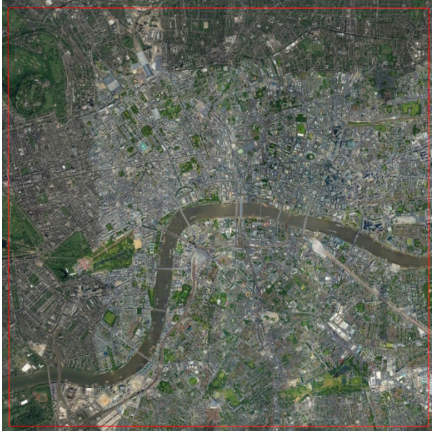 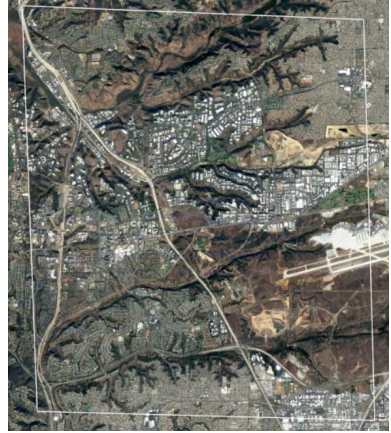

(a) London Dataset                                    (b) San Diego Dataset

Figure 6. London and San Diego datasets. Quadrilaterals in (a) and (b) are covering area of London and San Diego datasets.

### 5.2 Results and Analysis

In this paper, we define a group of correspondences as the image correspondences (at least two views) corresponding to a same object space point. In London dataset, our work matched 22640 groups of correspondences with 15433 groups being two-degree overlaps, 4761 being three-degree overlaps, 1839 groups being four-degree overlaps and the remaining 607 groups being five-degree overlaps. In San Diego Dataset, 15197 groups of correspondences were matched with 13795 groups being two-degree overlaps, 1019 groups being three-degree overlaps, 294 being four-degree overlaps and the remaining 89 groups being five-degree overlaps. We projected the object space points of each group onto all images, and measured the distances between the original image points and the corresponding projected points. We evaluated our bundle adjustment results by averaging the distance in x direction (avg_x), averaging the distance in y direction (avg_y), averaging the distance in image plane (avg_xy), selecting the maximum distance in x direction (max_x), selecting the maximum distance in y direction (max_y), selecting the maximum distance in image plane (max_xy), as shown in Table 2. "Before" means the repojection accuracy before bundle adjustment, and "After" means the reprojection accuracy after bundle adjustment. All the accuracy evaluation results only kept three decimal places.

Table 2. Bundle Adjustment Accuracy Evaluation (Unit: Pixel)

| Dataset | | avg_x | avg_y | avg_xy | max_x | max_y | max_xy |
|---|---|---|---|---|---|---|---|
| London | Before | 0.638 | 1.005 | 1.368 | 24.615 | 87.934 | 91.314 |
| | **After** | **0.190** | **0.114** | **0.243** | **15.938** | **54.445** | **56.730** |
| San Diego | Before | 1.365 | 3.315 | 3.718 | 13.066 | 17.991 | 18.892 |
| | **After** | **0.154** | **0.198** | **0.269** | **2.597** | **3.963** | **3.973** |

Table 2 shows that the our bundle adjustment method can compute sub-pixel accuracy bundle adjustment results with avg_x, avg_y below 0.2 pixels and avg_xy below 0.3 pixels, even though the satellite images came from different satellite cameras with different imaging date, viewing angles and resolutions. The max_x, max_y and max_xy were also improved a lot by 8.677 pixels to 34.584 pixels. However, the maximum reprojection errors were still large, especially in London datasets, it is because the correspondences groups with these maximum errors are mismatches. In future work, we will furtherly eliminate these mismatches, thus getting better bundle adjustment results.

To check the reprojection accuracy improvement for each individual image, we computed the reprojection errors for each pixel in image plane before/after the bundle adjustment, and presented these errors as dots with the sequence of point index, as shown in Figure 7 and Figure 8. To make more clear comparison, we manually eliminated the mismatches with reprojection errors larger than 10 pixels in London dataset. In both Figure 7 and Figure 8, the sub-figures in the first row and the second row are one-to-one corresponding, and the horizontal axis in each sub-figure represented the index of each point, the vertical axis represented the reprojection errors of each pixel with unit: pixel.

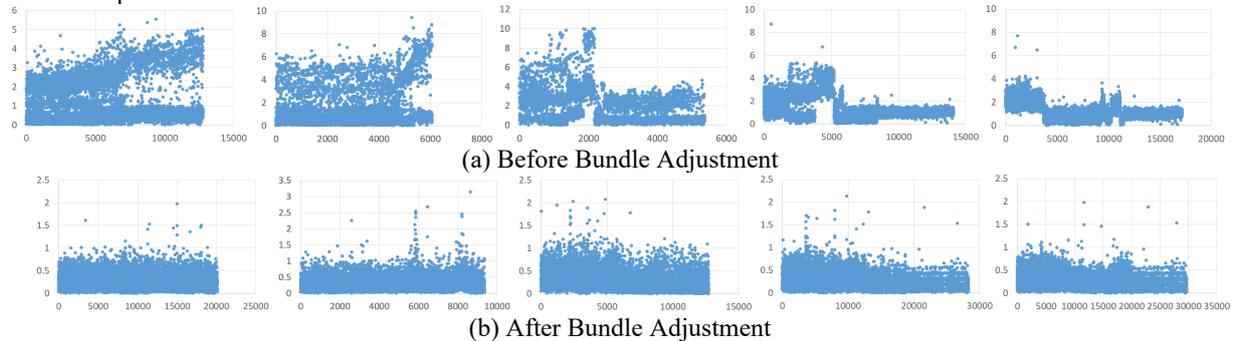

(a) Before Bundle Adjustment

(b) After Bundle Adjustment

Figure 7. Reprojection error comparison in London dataset. The five sub-figures in (a) and (b) respectively describes the reprojection error distributions in London datasets before/after bundle adjustment.

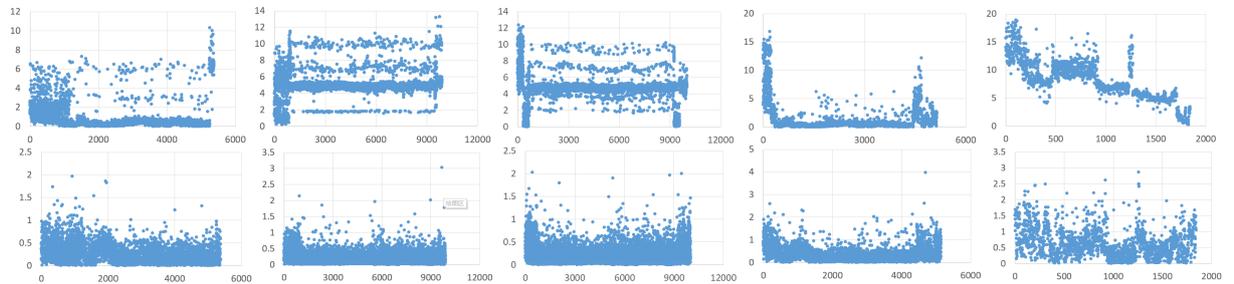

Figure 8. Reprojection error comparison in San Diego dataset. The five sub-figures in (a) and (b) respectively describes the reprojection error distributions in London datasets before/after bundle adjustment.

In Figure 7, before bundle adjustment, the average reprojection errors in image plane for each individual image are 1.271 pixels, 1.512 pixels, 1.917 pixels, 1.458 pixels and 1.142 pixels, and after bundle adjustment, these errors have been improved to 0.234 pixels, 0.326 pixels, 0.319 pixels, 0.225 pixels and 0.212 pixels. In Figure 8, the average reprojection errors before bundle adjustment are 0.873 pixels, 4.880 pixels, 4.660 pixels, 1.096 pixels and 7.974 pixels, and after bundle adjustment, these errors have been improved to 0.269 pixels, 0.207 pixels, 0.217 pixels, 0.361 pixels and 0.634 pixels. The comparison results showed that our method can compute sub-pixel bundle adjustment accuracies for each individual image. In San Diego dataset, the bundle adjustment accuracy (0.634 pixels) of the fifth image were not comparable with other images, it is because the matching points of the fifth image was much less than other images, thus its bundle adjustment result was easier to be influenced by some mismatches.

## 6. CONCLUSION

In this paper, we propose a multi-view, multi-source, large-scale bundle adjustment method for high-res satellite images, which can compensate geometric distortions and local linear radiometric distortions in satellite image datasets and compute sub-pixel accuracy bundle adjustment results. Our main work include: 1) a multi-view, multi-source matching method which compensates geometric distortions by plane rectification, uses epipolar constraint to reduce matching uncertainties and utilizes MBCensus to compensate local nonlinear radiometric distortions between image pairs; 2) an efficient solution to large-scale bundle adjustment scenario. During the adjustment computation, the maximum coefficient matrix is only $2N \times 2N$ with $N$ being the number of high-res satellite images.

Experiments on in-track satellite dataset and off-track satellite dataset (from different satellite cameras, different imaging dates and different GSD) demonstrated that our method is able to improve the reprojection errors from several pixels to sub-pixel accuracy. For in-track satellite dataset of London, the average reprojection errors in image planes have been improved from 1.368 pixels to 0.243 pixels, and for off-track satellite dataset of San Diego, the reprojection errors have been improved from 3.718 pixels to 0.269 pixels. We also observed that though bias correction techniques were used to eliminate mismatches, there are still several mismatches in the computation of bundle adjustment, thus influencing the final bundle adjustment results. In our future work, we plan to furtherly eliminate these mismatches and get sub-pixel accuracy bundle adjustment results in much larger scenarios (more than 1000 km$^2$).

## ACKNOWLEDGEMENT

The study is partially supported by the ONR grant (Award No. N000141712928). We would like to thank The Intelligence Advanced Research Projects Activity (IAPRA) for providing San Diego dataset.